\documentclass[11pt]{article}

\usepackage[preprint]{acl}
\usepackage{times}
\usepackage{latexsym}
\usepackage[T1]{fontenc}
\usepackage[utf8]{inputenc}
\usepackage{microtype}
\usepackage{inconsolata}
\usepackage{graphicx}

\usepackage{amsmath,amssymb}
\usepackage{booktabs}
\usepackage{array}
\usepackage{multirow}
\usepackage{xcolor}
\usepackage{enumitem}
\usepackage{listings}
\usepackage{xspace}
\usepackage{placeins}
\usepackage{tikz}
\usetikzlibrary{arrows.meta,positioning,fit}

\newcommand{\model}{\textsc{EdgeDeID}\xspace}
\newcommand{\fbeta}{F$_{2}$\xspace}
\newcommand{\PHI}{PHI\xspace}
\newcommand{\zeroleak}{zero-leak\xspace}
\newcommand{\ci}[2]{{\footnotesize[#1,\,#2]}}

\lstdefinestyle{xml}{
  basicstyle=\ttfamily\footnotesize,
  breaklines=true, columns=fullflexible,
  frame=single, framesep=4pt, rulecolor=\color{gray!40},
  backgroundcolor=\color{gray!5},
}

\title{\model: A 0.6B Clinical De-identifier Trained Without Real Patient
Data}

\author{%
  \textbf{Sina Daneshgar}\textsuperscript{1},
  \textbf{Raya Namazi}\textsuperscript{1},
  \textbf{Amirali Khosravi}\textsuperscript{2},
  \\
  \textbf{Pouria Rouzrokh}\textsuperscript{2,3},
  \textbf{Bardia Khosravi}\textsuperscript{2,3,\dag}
  \\[4pt]
  \textsuperscript{1}Sharif University of Technology \\
  \textsuperscript{2}Department of Radiology, Mayo Clinic \\
  \textsuperscript{3}Department of Radiology and Biomedical Imaging, Yale University
  \\[4pt]
  {\textsuperscript{\dag}Corresponding author:
  \href{mailto:bardia.khosravi@yale.edu}{bardia.khosravi@yale.edu}}
}

\begin{document}
\maketitle

\begin{abstract}
De-identifying clinical notes with a hosted language model discloses the information the step is meant to protect. We present \model, a 0.6B-parameter decoder-only de-identifier that runs locally and is trained without real patient text: three LLM teachers write synthetic notes, a single labeler re-annotates them into one convention, and the model learns to emit a typed XML schema of the identifiers it finds. On a 100-note synthetic in-domain benchmark with manually verified annotations it reaches $0.994$ micro-F1 and leaves no missed identifier in $99\%$ of notes, against $0.857$ and $12\%$ for the strongest of eight off-the-shelf systems. Quantized, it runs on CPU in 615 MB. Of three additions to supervised training, only a verifier adapter helps: it separates complete from incomplete outputs of other systems at $0.94$ AUROC, whereas reinforcement learning with a recall-weighted reward and a rule-based post-processing layer leave in-domain results unchanged. On MIMIC ICU nursing notes F1 falls to $0.48$. The errors follow two properties our synthetic notes lack, telegraphic style and low identifier density, and three hand-written in-context examples recover $0.08$ F1 without annotated target data.
\end{abstract}

\section{Introduction}\label{sec:intro}
The HIPAA Safe Harbor standard defines de-identification by enumeration: a
record counts as de-identified once eighteen categories of identifier have been
removed~\citep{hipaa_safeharbor}. Several have no counterpart in the
personally identifiable information (PII) taxonomies used for business text:
every element of a date except the year, ages over 89, geographic units smaller
than a state, medical record and health plan numbers, and device identifiers.
These are routine in clinical narrative, and a general-purpose PII filter has no
reason to flag them. Commercial privacy filters are described by their
developers as data-minimization components, not de-identification
systems~\citep{debourcy2026modelcardopenaiprivacy}, and in our evaluation the
strongest of them leaves at least one identifier in $88\%$ of our benchmark
notes, mostly in categories its schema has no label for.

Frontier language models de-identify clinical text
well~\citep{liu2025deidgptzeroshotmedicaltext},
but using one means sending unredacted records to a third party, which is the
disclosure de-identification exists to prevent, or first negotiating a business
associate agreement. A model small enough to run where the data lives avoids
both, but it makes more mistakes, and its two kinds of mistake differ in cost.
Over-redaction reduces the utility of a note, whereas a missed identifier is a
disclosure that cannot be undone, and even strong de-identifiers leave some
notes re-identifiable~\citep{morris2024diri}. Recall therefore matters more
than precision. We ask how far such a model can go without real training data
and without scaling up. Our contributions are:

\begin{itemize}[leftmargin=*,itemsep=2pt,topsep=3pt]
  \item \model, a 0.6B generative de-identifier trained on multi-teacher
  synthetic notes relabeled into one convention, without any real \PHI. On
  a $100$-note synthetic in-domain benchmark with manually verified annotations it
  reaches $0.994$ micro-F1 and leaves no missed \PHI in $99\%$ of notes, against $0.857$ and
  $12\%$ for the best of eight off-the-shelf systems; the same backbone without fine-tuning
  scores $0.199$. Quantized, it runs on a CPU in 615\,MB.
  \item A verifier adapter that judges whether a candidate annotation is
  complete, reaching $0.94$ AUROC on the outputs of nine other systems.
  Paired tests show that, in-domain, neither GRPO with a recall-weighted reward
  nor a rule-based detector layer improves a supervised model that already has
  $0.999$ recall, and we explain why a self-assessment tag trained by reward
  alone is never emitted.
  \item An analysis of transfer to MIMIC ICU nursing
  notes~\citep{neamatullah2008automated} that scores Safe Harbor identifiers
  separately. F1 drops to $0.478$, below an encoder trained on real i2b2 notes
  ($0.675$). Three quarters of the missed patient names and dates fall in notes
  with an empty or unparseable output, and the failures follow register
  (writing style) and \PHI density rather than clinical content. Three
  hand-written in-context examples raise F1 by $0.08$.
\end{itemize}

Code and the evaluation harness are available at
\url{https://github.com/BardiaKh/EdgeDeID}, and the synthetic corpus and model
weights at \url{https://huggingface.co/EdgeDeID}.

\section{Related Work}\label{sec:related}
\paragraph{Clinical de-identification.}
Anonymization of non-medical text has a large
literature~\citep{deusser2025survey}. Clinical work is mostly measured against
the i2b2/UTHealth shared tasks, which fixed a common corpus and a common set of
\PHI categories~\citep{stubbs2015overview,stubbs2015annotating}. Methods have
gone from hand-written rules~\citep{norgeot2020philter} to recurrent
taggers~\citep{dernoncourt2017deidentification} to fine-tuned
encoders~\citep{johnson2020deidentification,chambon2023automated,prakash2026improving},
which do well on notes like their training data and worse on notes from another
hospital or document type~\citep{lee2021transferability,Kuo2025}. Data
augmentation narrows that gap~\citep{yue2020phicon}, but systems that run
without human review still layer rules on top of the
model~\citep{kocaman2023accuracyautomateddeidentificationlarge}. Misses have
consequences: LLMs can re-identify patients from notes that were supposed to be
de-identified~\citep{morris2024diri}, sometimes through quasi-identifiers that
Safe Harbor never requires removing~\citep{jiang2026paradox}.

\paragraph{LLMs for de-identification.}
DeID-GPT showed that a frontier proprietary model can tag \PHI without
task-specific training~\citep{liu2025deidgptzeroshotmedicaltext}. On real
notes, zero-shot GPT-class models seldom flag text that is not \PHI but miss a
fair share of text that is~\citep{altalla2025evaluating}, and they trail
purpose-built systems~\citep{kocaman2025zeroshot}. Prompts that spell out an
institution's categories close much of that
gap~\citep{palacios2026institution}, and LLMs fine-tuned on annotated notes,
including the nursing-note corpus we evaluate on, beat encoder
baselines~\citep{panchal2026benchmarking}. Other work adds date
normalization~\citep{Dai2025}, deployable
pipelines~\citep{wu2025largelanguagemodelempowered,singh2025redactor}, or
adversarial anonymization~\citep{staab2025anonymizers}, and two studies
quantify what LLM redaction costs in lost clinical
information~\citep{aghakasiri2025doctororderedsurveyingllmbased} and in
inference cost~\citep{zambare2026fairefficientdeidentificationquantifying}.
Nearly all of this work relies on large models reached through an API.

\paragraph{Synthetic data and small models.}
Because real \PHI rarely leaves a hospital, several groups train on generated
text. \citet{vakili2025data} trained on text an LLM both wrote and annotated,
lost little relative to real data, and found that the annotating model mattered
most, which is why we relabel. \citet{kim2024generalizing} used GPT-4 to
augment redacted notes, and MedDeID, the closest to our work, trains a compact
encoder only on synthetic Dutch notes and approaches a model trained on
hospital data~\citep{hellemans2026meddeid}. Measures of drift between corpora
correlate moderately with how a synthetically trained tagger scores on real
text~\citep{cornelius2026measuring}. Small models are argued to be the realistic
route to clinical NLP when privacy and cost are
binding~\citep{garg2025slmhealthcare}: LLM anonymizers have been run entirely
on hospital
hardware~\citep{doi:10.1056/AIdbp2400537,Dormus2025,Hirsch2026.05.28.26353952},
and small taggers, often distilled from a larger
LLM~\citep{zhou2024universalner}, stay competitive on
premises~\citep{gunay2024llmsintheloop2expertsmall,posada2025shield,posada2026tide}.
General-purpose PII filters are also
compact~\citep{debourcy2026modelcardopenaiprivacy,zaratiana2024gliner,zaratiana2026gliner2pii}
but target web and enterprise text, and an independent evaluation found the
OpenAI Privacy Filter weak on medical text~\citep{uppala2026opf}, as do we.
\model differs from these systems in two ways: it is a generative model that
writes a typed schema instead of labeling spans, and it never sees real text
during training. Our results add a failure case to this literature, which we
trace to register and \PHI density.

\paragraph{Reinforcement learning and verification.}
Group Relative Policy Optimization (GRPO) compares each sampled completion with
the others in its group and needs no value
network~\citep{shao2024deepseekmath,guo2025deepseekr1}. After supervised
fine-tuning it has improved information
extraction~\citep{you2026larl,chen2026ner1} and clinical text
classification~\citep{wei2026reinforcement}, although analyses find that it
mostly sharpens behavior the base model already
produces~\citep{wu2025invisible,yue2025does} and that the KL penalty limits how
far the policy can move~\citep{mroueh2025rlvr}. We know of no prior application
to clinical de-identification. LLM judges are poor at spotting omissions in
clinical notes unless the task is broken down for
them~\citep{fox2026omission}; our verifier trains a dedicated adapter for this
instead.

\section{Method}\label{sec:method}
\paragraph{Task and output.}
A clinical note $x$ contains a set of \PHI spans $Y(x)=\{(t_i,s_i)\}$, where
$s_i$ is a substring of $x$ and $t_i$ one of ten types (\textsc{name},
\textsc{age}, \textsc{date}, \textsc{place}, \textsc{address},
\textsc{phone}, \textsc{email}, \textsc{id}, \textsc{url}, \textsc{device})
chosen to cover the direct identifiers of Safe
Harbor~\citep{hipaa_safeharbor}. The model $f_\theta$ writes $\hat{Y}$ as one
typed XML object, \texttt{<phi\_entities>}, in which a name carries its surface
variations, a date its original and normalized forms, and an id a subtype
(Appendix~\ref{app:schema}). The training reward, the QC layer, and the metric
all read this object, so an output is usable only if it parses. We evaluate
detection and leave masking and surrogate generation out of scope.

\begin{figure*}[t]
  \centering
  \includegraphics[width=\textwidth]{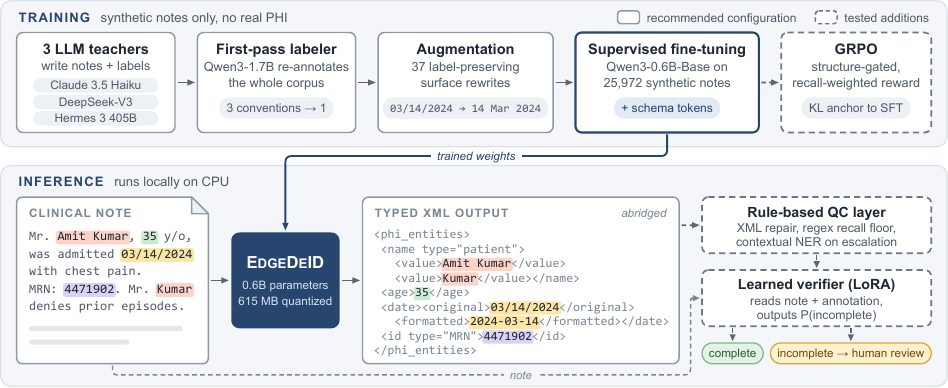}
  \caption{Overview of \model. The solid boxes are the configuration we
  recommend, which uses SFT alone. The dashed boxes are the three additions we
  tested on top of it: GRPO with its self-assessment tag
  (Section~\ref{sec:rewards}), and the rule-based QC layer and the learned
  verifier (Section~\ref{sec:reliability}).}
  \label{fig:pipeline}
\end{figure*}

Figure~\ref{fig:pipeline} gives an overview. Training builds a synthetic
corpus, fine-tunes a small model on it with supervised fine-tuning (SFT), and
optionally continues with GRPO. At inference, a rule-based QC layer and a
learned verifier can be placed after the model.

\subsection{Synthetic Data}\label{sec:data-method}
Each training note is written by one of three LLM teachers with distinct
styles, Claude 3.5 Haiku~\citep{anthropic2024claude35haiku}, DeepSeek-V3~\citep{deepseekai2024deepseekv3}, or Hermes~3
405B~\citep{teknium2024hermes3}, which receives a target
length and a list of \PHI to include and returns the note with its own
annotation. The teachers label the same spans differently and have their own
blind spots (Hermes~3, for example, tends to miss dates written as
\textit{DD-MM-YY}), so we do not train on their labels. Instead, we fine-tune
Qwen3-1.7B-Base~\citep{qwen3technicalreport} on a small set of teacher-written
notes whose labels were cleaned with rules and heuristics, and use it as a
\emph{first-pass labeler} that re-annotates the whole corpus, replacing three
conventions with one. After
dropping malformed rows, we apply 37 augmentations that change how an entity
is written but not what it refers to, such as the format of a date, the
spelling of a name, or the separators in a phone number. The final corpus has
$25{,}972$ training notes and a held-out synthetic split of $404$.

\subsection{Supervised Fine-Tuning}\label{sec:sft}
We fine-tune Qwen3-0.6B-Base~\citep{qwen3technicalreport} on a plain-text
template with three blocks, \texttt{<instructions>}, \texttt{<examples>}, and
\texttt{<note>}. The instructions describe the task and the schema. Worked
examples appear in $12\%$ of training prompts, so the model learns the task
both with and without them and can take few-shot examples at inference. SFT,
GRPO, and inference share this template, and the causal language-modeling loss
is applied only to the target XML.

\paragraph{Schema-aware tokenizer extension.}
The base tokenizer splits schema tags such as \texttt{</phi\_entities>} into
several sub-word pieces. We add $69$ tokens for the container, content, and
attribute tags of the schema and initialize each new embedding to the mean of
the existing ones. Over the training set this shortens a sample by $15.3\%$ and
the annotation, the only text the model generates, by $37.5\%$ ($536$ to $335$
tokens on average), removing more than a third of the decoding steps per
note.

\subsection{GRPO with a Structure-Gated Reward}\label{sec:rewards}
GRPO~\citep{shao2024deepseekmath} continues training from the SFT checkpoint.
For each prompt the policy samples $G=8$ completions, and each completion's
advantage is its reward standardized against the others for the same prompt.
The reward for a completion $c$ is
\begin{equation}
  R(c) =
  \begin{cases}
    0, & \text{if } c \text{ does not parse,}\\[2pt]
    \sum_{k} w_k\, r_k(c), & \text{otherwise,}
  \end{cases}
  \label{eq:reward}
\end{equation}
a weighted sum of four terms in $[0,1]$. \emph{Presence} ($w=0.30$) is the
fraction of tagged entities that appear verbatim in the note, which penalizes
invented \PHI. \emph{Completeness} ($w=0.40$) is entity-level \fbeta against
the reference. \emph{Calibration} ($w=0.15$) rewards the self-assessment tag
(Section~\ref{sec:selfassess}): it compares the model's claim,
\texttt{complete} or \texttt{incomplete}, with whether the annotation really is
complete, and a false claim of completeness and a missing tag both earn
zero. \emph{Length} ($w=0.15$) gives full credit up to
$1{,}024$ tokens. No term needs human labels or an LLM judge
(Appendix~\ref{app:reward}). A KL penalty to the SFT policy
($\beta_{\mathrm{KL}}=0.04$) anchors training, which runs for one pass over the
training set.

\subsection{Reliability Mechanisms}\label{sec:reliability}
A missed identifier is the costly error, so we test three ways of catching one
once the model has written its annotation. They differ in who judges the
annotation, when, and what happens to it (Table~\ref{tab:mechanisms}).

\begin{table}[t]
  \centering\small
  \setlength{\tabcolsep}{2pt}
  \begin{tabular}{@{}l >{\raggedright\arraybackslash}p{1.55cm}
                  >{\raggedright\arraybackslash}p{1.65cm}
                  >{\raggedright\arraybackslash}p{2.0cm}@{}}
    \toprule
     & Self-assess.\ tag & QC layer & Verifier \\
    \midrule
    Judge          & the extractor & rules and spaCy NER & LoRA adapter \\
    When           & same pass & after generation & second pass \\
    Training       & GRPO reward & none & SFT, then GRPO \\
    Result         & verdict & edited annotation & $P(\text{incomplete})$, review \\
    Other systems  & no & yes & yes \\
    \bottomrule
  \end{tabular}
  \caption{The three reliability mechanisms. ``Other systems'': whether the
  mechanism can be applied to the output of another de-identifier.}
  \label{tab:mechanisms}
\end{table}

\paragraph{Self-assessment tag.}\label{sec:selfassess}
The extractor judges its own annotation in the same generation pass, by writing
\texttt{<qc\_status>complete</qc\_status>} or \texttt{incomplete} after the
closing \texttt{</phi\_entities>}. The tag is absent from the supervised
targets, so the calibration term of the GRPO reward (Section~\ref{sec:rewards})
is its only training signal. A note the model marks incomplete is flagged for
review. The tag needs no extra model and no extra pass.

\paragraph{Rule-based quality-control (QC) layer.}\label{sec:qc}
A deterministic post-processor edits the annotation in three steps.
\emph{Repair} turns malformed XML into a well-formed annotation, falling back
to a recovering parser that keeps every tag it can read. \emph{Pattern
detection} runs regular expressions over the note for the six types with a
fixed surface form (email, phone, URL, id, age, and date, accepting only
calendar-valid numeric dates) and adds any match the model left untagged, so
recall on these types cannot fall below that of the patterns. The \emph{NER
fallback} runs spaCy to find names, places, and addresses, skipping
abbreviations, generic institution words, and mostly upper-case notes. Because
it is slow and imprecise, it runs only on notes that look unreliable: those
whose XML could not be repaired, in which pattern detection added three or
more spans, or that the self-assessment tag marks incomplete
(Appendix~\ref{app:qc_ablation}). The layer needs no training, only adds spans,
and can be applied to any system's output.

\paragraph{Learned verifier.}\label{sec:verifier}
The verifier is a second pass that judges an annotation without editing it. A
LoRA adapter~\citep{hu2022lora} on the SFT model reads a note and a
\emph{candidate} annotation from any system and returns a verdict in the same
format as the self-assessment tag. The two verdicts differ at their first
token, so one forward pass gives $P(\texttt{incomplete})$, and annotations
above a threshold go to human review. The verifier thus moves self-assessment
out of the extractor's own generation, where the tag must be learned at the end
of a long output, into a pass whose only output is the verdict. The adapter is
switched off during extraction, so the extractor is unchanged. Its training
candidates come only from synthetic training notes: gold annotations corrupted
by dropping entities, truncating spans, deleting categories, or adding
spurious entities, together with the outputs of four baseline systems
(Appendix~\ref{app:hparams}). Each candidate is labeled by the \zeroleak rule,
and the adapter is trained with SFT and then with GRPO on its verdict.

\section{Experimental Setup}\label{sec:setup}
\subsection{Evaluation Data}\label{sec:data}
\paragraph{In-domain benchmark.}
Our primary test set is $100$ synthetic clinical notes, generated separately
from the training corpus, whose annotations were manually verified and
corrected so that every gold value appears verbatim in its note. They match the style of the
training data, so we treat results on them as an upper bound.

\paragraph{MIMIC ICU notes.}
For out-of-domain evaluation we use the PhysioNet de-identification
corpus~\citep{neamatullah2008automated,goldberger2000physiobank}: $2{,}434$
ICU nursing notes whose original identifiers were replaced with realistic
surrogates, with $1{,}778$ human-annotated \PHI spans. We score against this
human annotation, not against the output of the
rule-based de-identifier distributed with it (Appendix~\ref{app:mimic}). About
$60\%$ of the notes contain no \PHI, so we do not report \zeroleak here. The
annotators used a broader definition of \PHI than HIPAA, which adds clinician
names and years~\citep{neamatullah2008automated}. By our count only $823$ of
the gold spans are Safe Harbor identifiers of a patient or relative, and $593$
are clinician names (Appendix~\ref{app:mimic}), so we also report recall on the
Safe Harbor spans alone.

\subsection{Baselines}\label{sec:baselines}
We run eight local off-the-shelf systems as a site without annotated data
would, with the developers' default configurations and labels mapped to our
types (Appendix~\ref{app:baselines}): the pattern and NER tools
Presidio~\citep{presidio} and spaCy~\citep{honnibal2020spacy}, the
open-vocabulary tagger GLiNER~\citep{zaratiana2024gliner}, the clinical encoders
Stanford-deid~\citep{chambon2023automated}, RoBERTa-i2b2 and
BERT-i2b2~\citep{robustdeid}, and OpenMed-PII~\citep{panahi2025openmed}, and
the general-purpose OpenAI Privacy Filter
(OPF)~\citep{debourcy2026modelcardopenaiprivacy}. Zero-shot Qwen3-0.6B-Base
with our prompt is a lower bound at the same model size.

\subsection{Metrics and Implementation}\label{sec:metrics}
We match typed entities by case-insensitive string comparison within each note.
A gold entity counts as found only when a prediction of the same type contains
all of it, since a partly covered identifier still leaks. Precision is more
lenient: a prediction is correct if it overlaps a gold entity in either
direction. We report micro-averaged precision, recall, and F1; macro-F1 over
the categories each system can produce; \fbeta, the F-score with $\beta=2$,
which weights recall above precision; the \zeroleak rate, the share of notes
with no missed gold entity; and the XML parse rate. Our own configurations are
compared with a paired bootstrap over notes ($1{,}000$ resamples, $95\%$
intervals). All \model results use greedy decoding that stops at the closing
\texttt{</phi\_entities>} tag. Each configuration was trained once with a fixed
seed, so the intervals reflect resampling of the test set, not training
variance. Appendix~\ref{app:hparams} lists hyperparameters and compute.

\section{Results}\label{sec:results}

\subsection{In-Domain Benchmark}\label{sec:main-results}
\begin{table*}[t]
  \centering\small
  \setlength{\tabcolsep}{5pt}
  \begin{tabular}{l r c c c c c c c}
    \toprule
    System & Params (M) & P & R & F1 & \fbeta & Macro-F1 & Zero-leak & Parse \\
    \midrule
    OPF                      & 1{,}500$^\dagger$ & 0.983 & 0.760 & 0.857 & 0.796 & 0.852 & 0.12 & 1.00 \\
    GLiNER                   & 195.2 & 0.916 & 0.681 & 0.782 & 0.718 & 0.617 & 0.05 & 0.99 \\
    Presidio                 & --    & 0.764 & 0.569 & 0.652 & 0.600 & 0.545 & 0.05 & 1.00 \\
    Stanford-deid            & 108.9 & 0.646 & 0.554 & 0.597 & 0.570 & 0.366 & 0.02 & 1.00 \\
    spaCy                    & --    & 0.764 & 0.459 & 0.573 & 0.499 & 0.514 & 0.02 & 0.99 \\
    OpenMed-PII              & 434.1 & 0.672 & 0.345 & 0.456 & 0.382 & 0.390 & 0.00 & 1.00 \\
    RoBERTa-i2b2             & 354.4 & 0.785 & 0.299 & 0.433 & 0.341 & 0.412 & 0.00 & 1.00 \\
    BERT-i2b2                & 107.8 & 0.850 & 0.224 & 0.355 & 0.263 & 0.203 & 0.00 & 1.00 \\
    Qwen3-0.6B-Base (0-shot) & 596.0 & 0.586 & 0.120 & 0.199 & 0.142 & 0.132 & 0.02 & 0.27 \\
    \midrule
    \model (SFT)                              & 596.0 & \textbf{0.990} & \textbf{0.999} & \textbf{0.994} & \textbf{0.997} & \textbf{0.986} & \textbf{0.99} & 1.00 \\
    \quad + QC                                & 596.0 & 0.989 & \textbf{0.999} & \textbf{0.994} & \textbf{0.997} & 0.985 & \textbf{0.99} & 1.00 \\
    \model (GRPO)                          & 596.0 & \textbf{0.990} & \textbf{0.999} & \textbf{0.994} & \textbf{0.997} & \textbf{0.986} & \textbf{0.99} & 1.00 \\
    \quad + QC                                & 596.0 & 0.989 & \textbf{0.999} & \textbf{0.994} & \textbf{0.997} & 0.985 & \textbf{0.99} & 1.00 \\
    \bottomrule
  \end{tabular}
  \caption{Results on the 100-note in-domain benchmark. External systems are
  sorted by F1. Every system is scored with the protocol of
  Section~\ref{sec:metrics}, and a category a system cannot produce still
  counts against its recall and \zeroleak rate. Parse is the share of outputs
  that parse after conversion to our schema. Bold marks the best value in each
  column. $^\dagger$Mixture of experts with about 50M active parameters.}
  \label{tab:main}
\end{table*}

Table~\ref{tab:main} compares all systems on the benchmark. The supervised
model reaches $0.994$ F1 and $0.999$ recall and misses no \PHI in $99$ of the
$100$ notes. The best external system, the 1.5B-parameter OPF, reaches $0.857$
F1 with a \zeroleak rate of $0.12$, and no external system has recall above
$0.76$. Zero-shot, the same base model scores $0.199$ F1 and produces
parseable output for only $27\%$ of notes, so training accounts for both the
accuracy and the output format. \model scores at least $0.92$ F1 on every type
(Appendix~\ref{app:per-entity}). The external systems do well on types with a
regular surface form, such as emails, and poorly on types that depend on
context; OPF's eight categories~\citep{debourcy2026modelcardopenaiprivacy}
cover no age, place, or device, so every note containing one is a leak. The
comparison favors \model, because the benchmark is in its domain and in no
baseline's. Section~\ref{sec:mimic} tests the reverse.

\subsection{Additions to the Supervised Model}\label{sec:ablation}
On the benchmark, the supervised model already reaches $0.999$ recall and
$0.990$ precision, which leaves little room for improvement. Of the three
additions we test, only the learned verifier contributes a capability that the
supervised model lacks (Table~\ref{tab:verifier}).

\paragraph{GRPO.}
GRPO left the benchmark predictions unchanged, so all paired differences from
SFT are zero (lower block of Table~\ref{tab:main};
Appendix~\ref{app:significance}). The two models differ only on the MIMIC
nursing notes (Section~\ref{sec:mimic}).

\paragraph{Self-assessment tag.}
Neither the SFT nor the GRPO model produced the tag in any output, including
when the GRPO model was prompted with its own training instructions, which
make the tag mandatory. The SFT targets end at \texttt{</phi\_entities>}, so
almost no sampled completion contained the tag, and a term that is zero for
every completion in a group contributes nothing to the group-standardized
advantage. The calibration term therefore had no effect during training
(Appendix~\ref{app:qc_calibration}).

\paragraph{QC layer.}
The QC layer brought no measurable benefit. Every output already parsed, no
note met the criteria for escalation to the NER pass, and pattern detection
added a single entity, which was not \PHI. Precision changed by $-0.001$, and
all other metrics were unchanged (Appendix~\ref{app:qc_ablation}). Because SFT
and GRPO produce identical annotations on the benchmark, these results apply
to both models.

\begin{table}[t]
  \centering\small
  \setlength{\tabcolsep}{4pt}
  \begin{tabular}{@{}l c c c@{}}
    \toprule
    Verifier & Synth.\ val & Benchmark & MIMIC \\
    \midrule
    Zero-shot (no adapter) & 0.537 & 0.600 & 0.658 \\
    SFT adapter            & \textbf{0.813} & 0.987 & \textbf{0.661} \\
    SFT + GRPO adapter     & 0.809 & \textbf{0.989} & \textbf{0.661} \\
    \bottomrule
  \end{tabular}
  \caption{AUROC of the verifier for detecting incomplete annotations. Synth.\
  val: $800$ held-out synthetic candidates. Benchmark: outputs of the $11$
  systems of Table~\ref{tab:main} on the $100$ notes, $80\%$ of them
  incomplete. MIMIC: outputs of $11$ systems on $2{,}434$ notes, $18\%$ of them
  incomplete.}
  \label{tab:verifier}
\end{table}

\paragraph{Verifier.}\label{sec:verifier-results}
Trained only on synthetic candidates, the SFT adapter separates complete from
incomplete outputs of the eleven systems on the benchmark at $0.987$ AUROC
(Table~\ref{tab:verifier}). Most complete outputs in this pool are \model's
own, and on the nine other systems alone the AUROC is $0.938$. The hardest
case is OPF, whose incomplete annotations miss whole categories while
keeping near-perfect precision: on its outputs alone the adapter reaches
$0.789$, below the base model without an adapter ($0.819$). At the
reward-optimal threshold ($P(\texttt{incomplete})>1/3$) the adapter flags
$75\%$ of the outputs, which contain $93\%$ of the incomplete ones
(Appendix~\ref{app:verifier}). A GRPO stage on the verdict leaves AUROC
unchanged and worsens calibration (ECE $0.139 \rightarrow 0.213$). Because
\model's own benchmark outputs include only one incomplete note in $100$, the
benchmark cannot measure the verifier's value as a self-check.

\subsection{Out-of-Domain: MIMIC ICU Notes}\label{sec:mimic}
\begin{table}[t]
  \centering\footnotesize
  \setlength{\tabcolsep}{3pt}
  \begin{tabular}{@{}l c c c c r@{}}
    \toprule
    System & P & R & F1 & SH-R & Leak \\
    \midrule
    RoBERTa-i2b2        & 0.597 & \textbf{0.777} & \textbf{0.675} & \textbf{0.854} & \textbf{87} \\
    Stanford-deid       & 0.445 & \underline{0.654} & 0.529 & 0.362 & 281 \\
    BERT-i2b2           & \textbf{0.616} & 0.458 & 0.525 & 0.266 & 314 \\
    GLiNER              & 0.387 & 0.614 & 0.475 & 0.599 & 201 \\
    OPF                 & 0.539 & 0.331 & 0.410 & 0.298 & 314 \\
    OpenMed-PII         & 0.246 & 0.564 & 0.342 & \underline{0.756} & \underline{138} \\
    Presidio            & 0.065 & 0.514 & 0.115 & 0.574 & 222 \\
    spaCy               & 0.034 & 0.530 & 0.063 & 0.514 & 236 \\
    Qwen3-0.6B (0-shot) & 0.007 & 0.042 & 0.012 & 0.046 & 375 \\
    \midrule
    \model (SFT)               & 0.407 & 0.580 & 0.478 & 0.666 & 160 \\
    \quad + QC                 & 0.280 & 0.626 & 0.387 & 0.729 & 148 \\
    \quad few-shot ($k{=}3$)   & 0.563 & 0.561 & 0.562 & 0.622 & 181 \\
    \model (GRPO)              & 0.518 & 0.528 & 0.523 & 0.598 & 195 \\
    \quad few-shot ($k{=}3$)   & \underline{0.606} & 0.607 & 0.607 & 0.663 & 170 \\
    \quad few-shot + QC        & 0.597 & 0.621 & \underline{0.609} & 0.699 & 164 \\
    \bottomrule
  \end{tabular}
  \caption{MIMIC ICU nursing notes ($2{,}434$ notes) scored against the human
  gold standard. External systems are sorted by F1 and followed by our
  configurations, which are zero-shot unless marked few-shot (three
  hand-written example notes in the prompt); + QC adds the rule-based layer.
  SH-R is recall on the $823$ Safe Harbor spans, where a span counts as caught
  if a prediction of any type covers it. Leak is the number of the $399$ notes
  containing Safe Harbor \PHI in which at least one such span is missed (lower
  is better). Best value per column in bold, second best underlined.}
  \label{tab:mimic}
\end{table}

Table~\ref{tab:mimic} applies the same protocol to MIMIC ICU nursing notes,
which do not resemble our synthetic notes. Here the SFT model reaches $0.478$
F1, against $0.675$ for RoBERTa-i2b2, which was trained on real i2b2 notes much
closer to this distribution. Together with Table~\ref{tab:main}, this shows
that each system performs best on text resembling its training data and that
none is robust across both registers. On the Safe Harbor spans RoBERTa-i2b2
remains first, and the QC layer raises SFT's Safe Harbor recall from $0.666$
to $0.729$, at a cost in precision.

\paragraph{Most misses are whole-note failures.}
Of the $713$ patient and relative names and dates, the SFT model misses $200$.
For $155$ of these, the output for the whole note was empty ($103$) or
unparseable ($52$), which never occurs on the benchmark; in notes with a
well-formed annotation the miss rate is $8.1\%$ (Table~\ref{tab:miss}). The
training notes, like the benchmark, are dense in \PHI ($10.8$ mentions per
$1{,}000$ characters) and introduce it explicitly. MIMIC notes carry $0.92$,
$30\%$ of them are mostly upper case, and $81\%$ of their dates are a bare
month and day. Small controlled edits ($n{=}25$ each) are consistent with this
account: prepending a ``Patient: \emph{name}'' header recovers $76\%$ of missed
names and keeping only the sentence that contains the target recovers $68\%$,
whereas title-casing the name recovers $20\%$. Register noise applied to
benchmark notes costs far less (Appendix~\ref{app:robustness}).

\paragraph{False positives are register errors.}
The model reads clinical shorthand as identifiers. It tags ``Aline'' (an
arterial line), ``Swan'' (a Swan-Ganz catheter), and ``Foley'' as names, and it
treats shift boundaries (``7a-7p''), military times (``2030''), and numeric
settings (``10/5'') as dates. Differences in annotation convention explain
little of the loss: ignoring entity type and discarding predicted ages under
$90$, which MIMIC does not annotate, raises SFT precision only from $0.407$ to
$0.454$.

\paragraph{In-context examples recover part of the gap.}
Because the template accepts worked examples (Section~\ref{sec:sft}), part of
the register gap can be addressed at inference time without retraining. With
three short hand-written example notes in the prompt, F1 rises by $0.083$ for
SFT ($0.478 \rightarrow 0.562$) and by $0.084$ for GRPO ($0.523 \rightarrow
0.607$), both significant (Appendix~\ref{app:significance}); with the QC layer, GRPO
reaches $0.609$, second only to RoBERTa-i2b2. For SFT the gain is in
precision ($0.407 \rightarrow 0.563$). GRPO gains in both precision, reaching
$0.606$, above RoBERTa-i2b2 ($0.597$), and recall ($0.528 \rightarrow 0.607$),
consistent with the examples correcting the register errors above. Safe Harbor
recall rises for GRPO ($0.598 \rightarrow 0.663$) but falls for SFT
($0.666 \rightarrow 0.622$), which returns an empty annotation for more notes
when given the examples.

\paragraph{GRPO and the reliability mechanisms on ICU notes.}
GRPO trades recall for precision on MIMIC ($-0.052$ and $+0.112$, both
significant; Appendix~\ref{app:significance}), which raises F1 to $0.523$ but
leaves more Safe Harbor spans uncovered. The self-assessment tag does not
appear here either. The QC layer moves the other way. On
SFT it raises Safe Harbor recall from $0.666$ to $0.729$ and reduces the notes
with a missed Safe Harbor span from $160$ to $148$ (for GRPO, $0.598$ to $0.650$
and $195$ to $185$), but SFT precision falls from $0.407$ to $0.280$: the layer
adds $1{,}376$ entities, of which $141$ match a gold span. Nearly all of them
come from the repair step, which recovers the tags of the $2.6\%$ of outputs
that do not parse; pattern detection adds $33$ entities, and the NER fallback
runs on one note. The verifier reaches an AUROC of $0.661$, no better than the
base model without an adapter ($0.658$; Table~\ref{tab:verifier}), since it
shares the extractor's backbone and training register. Unlike in-context
examples, neither mechanism corrects the register errors themselves.

\subsection{Efficiency}\label{sec:efficiency}
\begin{table}[t]
  \centering\small
  \setlength{\tabcolsep}{3pt}
  \begin{tabular}{@{}l r r r@{}}
    \toprule
    System & GPU & CPU & RSS \\
    \midrule
    Stanford-deid & 8    & 124      & 1{,}222 \\
    BERT-i2b2     & 9    & 119      & 1{,}233 \\
    RoBERTa-i2b2  & 14   & 335      & 2{,}118 \\
    GLiNER        & 18   & 192      & 2{,}418 \\
    OpenMed-PII   & 26   & 476      & 2{,}285 \\
    OPF           & 24   & 12{,}413 & 4{,}778 \\
    Presidio      & --   & 51       & 1{,}536 \\
    \midrule
    Ours, PyTorch            & 3{,}990 & 29{,}484 & 4{,}248 \\
    Ours, vLLM (bf16)        & 511  & --       & --      \\
    Ours, llama.cpp (Q4)     & --   & 3{,}931  & 615     \\
    \bottomrule
  \end{tabular}
  \caption{Median latency in ms per note ($20$ benchmark notes, processed one at
  a time after warm-up) on an NVIDIA L40S GPU and on $8$ threads of an AMD EPYC
  9354 CPU, with peak resident memory in MB for the CPU run. ``Ours'' is
  \model (SFT); baselines run in PyTorch. Q4 is 4-bit (Q4\_K\_M) GGUF
  quantization. The CPU host was shared, so CPU figures are conservative.}
  \label{tab:efficiency}
\end{table}

Table~\ref{tab:efficiency} reports single-note latency. Because it writes its
output token by token, \model is one to two orders of magnitude slower than the
encoder taggers, even though the schema tokens shorten each annotation by more
than a third (Section~\ref{sec:sft}). Served with vLLM~\citep{kwon2023efficient}
on one GPU, it takes $0.5$\,s per note and processes $31$ notes per second under
$100$ concurrent requests. Quantized to 4 bits and run with llama.cpp on eight
CPU threads, it takes $3.9$\,s per note and $615$\,MB of memory, the smallest
footprint in the table, against $12.4$\,s for OPF on the same CPU.
Quantization barely changes accuracy: the 4-bit model scores $0.994$ F1 with a
\zeroleak rate of $0.98$, against $0.99$ at full precision
(Appendix~\ref{app:latency}).

\section{Discussion}\label{sec:discussion}
\paragraph{Small models trained on synthetic data.}
General-purpose PII filters are not a substitute for a model trained on the
Safe Harbor categories. The strongest, a 1.5B model with near-perfect
precision, left an identifier in $88\%$ of benchmark notes, mostly in
categories it has no label for. Fine-tuning a 0.6B model on synthetic notes
alone closed that gap on the in-domain benchmark without real \PHI at any stage,
and the result runs on hardware an institution already owns. We see this as a
proof of concept: multi-teacher generation, consistent relabeling, and a typed
output schema could extend to other note types, languages, or identifier
definitions by changing the generator prompts and the schema.

\paragraph{Register and \PHI density.}
Our error analysis attributes the drop on MIMIC to the form of the notes rather
than their clinical content, and most missed names and dates fall in notes that
received no usable annotation. Hand-written in-context examples and
context-restoring edits both recover part of the loss, and both point to the
same remedy at training time: synthetic notes across a wider range of registers
and \PHI densities, including notes that contain no \PHI.

\paragraph{Reinforcement learning.}
In-domain, nearly every sampled completion received the same reward, and
GRPO left the model unchanged; on MIMIC, where there was room to improve, it
moved the model toward precision. This is consistent with reports that
reinforcement learning mainly amplifies behavior the base policy already
produces~\citep{wu2025invisible,yue2025does} and that a 0.5B base model does
not acquire the reasoning behavior that larger models learn through the same
procedure~\citep{tinyzero}. At this scale it adjusts how the model answers more
than what it can detect.

\paragraph{Three reliability mechanisms.}
A self-assessment tag that the extractor must learn to write at the end of its
own output never appeared, because a reward on a behavior the policy never
samples gives GRPO nothing to amplify. Posed instead as a second pass whose
only output is the verdict, self-assessment became learnable ($0.94$ AUROC on
other systems' in-domain outputs), although it did not transfer to MIMIC. The
rule-based QC layer is best treated as an optional high-recall mode for sites
that review the output: it adds nothing in-domain, but on MIMIC it raised Safe
Harbor recall from $0.666$ to $0.729$, mostly by recovering malformed outputs,
at a cost of $0.13$ in precision.

\section{Conclusion}\label{sec:conclusion}
We presented \model, a 0.6B generative de-identifier trained only on
synthetic clinical notes. General-purpose PII detectors, including a 1.5B
privacy filter, miss identifiers that clinical de-identification must remove,
whereas fine-tuning a small model on synthetic data reaches $0.994$ F1 on
in-domain notes and runs locally on a CPU in $615$\,MB. On MIMIC ICU
notes performance drops, which we trace to telegraphic register and low \PHI
density, and a few hand-written in-context examples recover part of the loss
without annotated data. Reinforcement learning added little to a supervised
model already near its ceiling. These results support synthetic-data
fine-tuning of small models as a practical route to local de-identification and
identify the registers and \PHI densities covered by synthetic data as the main
target for future work.

\section*{Limitations}
\paragraph{Training data and models.}
\model is trained only on synthetic notes from three LLM teachers, so its
coverage is bounded by what those teachers write, and its performance drops
on telegraphic, \PHI-sparse text such as MIMIC ICU notes
(Section~\ref{sec:mimic}). We used one backbone family at one size, together
with the base and teacher models available at the time of the study; newer or
larger models may behave differently. We did not ablate the pseudo-labeling
step or the tokenizer extension, and we did not train an encoder tagger on
our synthetic corpus, so we cannot say whether a generative model is the best
use of this data. An encoder would be faster at inference.

\paragraph{Reinforcement learning.}
Our GRPO results come from one reward design applied to a strong supervised
model, with one run per setting. At 0.6B parameters, reinforcement learning
may mainly adjust output style and format rather than underlying
capability~\citep{tinyzero,yue2025does}, so these findings may not hold for
larger models.

\paragraph{Evaluation.}
Our in-domain benchmark is small ($100$ notes), synthetic, and saturated for
our models. The i2b2/UTHealth corpora, the standard benchmark for this task,
were not accessible to us, so we could not report results on them. MIMIC ICU
notes are therefore our only real-text evaluation; they cover one note type
from one institution, exercise seven of our ten categories, and are closer to
the training data of the i2b2 baselines than to ours. Our Safe Harbor
classification of their gold spans is rule-based and is not a legal
determination. Prior work often reports token- or character-level scores above
$98\%$~\citep{dernoncourt2017deidentification,norgeot2020philter},
but most characters in a clinical note are not \PHI, so such scores stay high
even when whole identifiers are missed. We instead score entities under a
recall-first, per-note protocol that counts an identifier as protected only
when it is fully covered; this is closer to how a de-identifier is used, but
our numbers are not directly comparable with those reported earlier. We wrote
the QC rules while examining errors on our evaluation data, including MIMIC,
and a site would adjust them to its own notes. Each configuration is a single
training run, all work is in English, the controlled-edit experiments use $25$
examples each, and latency was measured on a shared server rather than a
consumer device.

\paragraph{Scope of the privacy claim.}
\zeroleak measures residual \PHI against a reference annotation. It does not
guarantee protection against re-identification, since a note may still be
re-identifiable from quasi-identifiers~\citep{morris2024diri}. We evaluate
detection only and do not generate surrogates.

\section*{Ethical Considerations}
Training uses only synthetic notes written by language models, and no patient
record was used for training or for the in-domain benchmark. The only clinical
corpus we use is the PhysioNet de-identification
corpus of MIMIC ICU notes~\citep{neamatullah2008automated}, in which the
original identifiers were replaced by surrogates before release. We obtained
it under the PhysioNet data use agreement, processed it only on local
hardware, never sent it to an external API, and do not redistribute it.

\model reduces the manual effort of de-identification, but it does not
guarantee anonymity and does not replace the legal and institutional
safeguards that govern the release of clinical data. Output intended for
release should be reviewed by a person and, where required, undergo expert
determination~\citep{hipaa_safeharbor}, and the model should be validated on
an institution's own notes before use. Because it runs locally, sensitive text
need not leave the institution.

\FloatBarrier
\bibliography{references}

\FloatBarrier
\appendix
\renewcommand{\topfraction}{0.9}
\renewcommand{\bottomfraction}{0.8}
\renewcommand{\textfraction}{0.07}
\renewcommand{\floatpagefraction}{0.8}
\renewcommand{\dbltopfraction}{0.9}
\renewcommand{\dblfloatpagefraction}{0.8}
\setcounter{topnumber}{4}
\setcounter{bottomnumber}{3}
\setcounter{totalnumber}{6}
\setcounter{dbltopnumber}{3}

\section{Output Schema}\label{app:schema}
Figure~\ref{fig:schema} shows an annotation in the output schema. Empty
categories are written as self-closing tags, so every output has the same ten
containers.

\begin{figure}[!htb]
\begin{lstlisting}[style=xml]
<phi_entities>
<names>
<name id="1" type="patient">
<variations>
<value>Amit Kumar</value>
<value>Kumar</value>
</variations>
</name>
</names>
<ages><age>35</age></ages>
<dates>
<date>
<original>Oct 1</original>
<formatted>????-10-01</formatted>
</date>
</dates>
<places>
<place id="1" type="hospital">NYU Langone Medical Center</place>
</places>
<addresses/> <phones/> <emails/>
<ids><id type="MRN">4471902</id></ids>
<urls/> <devices/>
</phi_entities>
\end{lstlisting}
\caption{An annotation in the \texttt{<phi\_entities>} schema (illustrative
values, whitespace condensed).}
\label{fig:schema}
\end{figure}

\section{Training Details}\label{app:hparams}
Tables~\ref{tab:hparams}, \ref{tab:grpo-hparams}, and~\ref{tab:ver-hparams}
list the configurations of the supervised stage, the GRPO stage, and the
verifier. All training ran on NVIDIA L40S GPUs: the first-pass labeler for
about $1$ hour on two GPUs, SFT for about $3$ hours on two GPUs, GRPO for
$59.6$ hours on one GPU, and the verifier adapter for $0.8$ hours (SFT stage)
and $2.2$ hours (GRPO stage) on one GPU. Inference benchmarks used one L40S.

\begin{table}[!htb]
  \centering\small
  \setlength{\tabcolsep}{4pt}
  \begin{tabular}{l p{3.6cm}}
    \toprule
    Hyperparameter & Value \\
    \midrule
    Base model            & Qwen3-0.6B-Base (labeler: Qwen3-1.7B-Base) \\
    Precision             & bfloat16 \\
    Epochs                & 1 \\
    Per-device batch size & 2 \\
    Gradient accumulation & 8 \\
    Optimizer             & AdamW \\
    Learning rate         & $5\times10^{-5}$ \\
    LR schedule           & warmup / constant / decay (5\% / 90\% / 5\%) \\
    Max sequence length   & 8{,}192 \\
    Loss                  & completion-only causal LM \\
    Seed                  & 2025 \\
    Hardware (SFT)        & 2 $\times$ NVIDIA L40S, ${\sim}3$ h \\
    Hardware (labeler)    & 2 $\times$ NVIDIA L40S, ${\sim}1$ h \\
    \bottomrule
  \end{tabular}
  \caption{Supervised fine-tuning.}
  \label{tab:hparams}
\end{table}

\begin{table}[!htb]
  \centering\small
  \setlength{\tabcolsep}{4pt}
  \begin{tabular}{l p{3.4cm}}
    \toprule
    Hyperparameter & Value \\
    \midrule
    Initialization               & SFT checkpoint, full fine-tuning (no LoRA) \\
    Precision                    & bfloat16 \\
    Learning rate                & $1\times10^{-5}$ \\
    Prompts per step             & 1 \\
    Completions per prompt ($G$) & 8 \\
    Per-device batch size        & 1 \\
    Gradient accumulation        & 8 \\
    Sampling temperature         & 1.0 \\
    Max completion length        & 1{,}024 tokens \\
    Reward weights               & presence 0.30, completeness 0.40, calibration 0.15, length 0.15 \\
    KL coefficient               & 0.04 \\
    Training length              & one pass over the 25{,}972 training notes (25{,}972 steps) \\
    Seed                         & 2025 \\
    Hardware                     & 1 $\times$ NVIDIA L40S, $59.6$ h \\
    \bottomrule
  \end{tabular}
  \caption{GRPO.}
  \label{tab:grpo-hparams}
\end{table}

\begin{table}[!htb]
  \centering\small
  \setlength{\tabcolsep}{4pt}
  \begin{tabular}{l p{3.6cm}}
    \toprule
    Hyperparameter & Value \\
    \midrule
    Base                  & \model SFT checkpoint (frozen) \\
    Adapter               & LoRA, rank 16, $\alpha$ 32, dropout 0.05 \\
    Training data         & $8{,}000$ synthetic training notes; corrupted gold annotations and outputs of BERT-i2b2, GLiNER, Presidio, and RoBERTa-i2b2 \\
    SFT stage             & $6{,}655$ candidates, 1 epoch, lr $1\times10^{-4}$, batch $2\times8$ \\
    SFT label balance     & 26\% incomplete \\
    GRPO stage            & $2{,}142$ candidates, 300 steps, lr $5\times10^{-6}$, $G=8$, gradient accumulation 4, temperature 1.0, no KL term \\
    GRPO label balance    & 45\% incomplete \\
    GRPO reward           & well-formed $+0.2$; correct $+0.8$; over-cautious $+0.4$; overconfident $0$ \\
    Hardware              & 1 $\times$ NVIDIA L40S; $0.8$ h (SFT), $2.2$ h (GRPO) \\
    \bottomrule
  \end{tabular}
  \caption{Verifier adapter.}
  \label{tab:ver-hparams}
\end{table}

\section{Reward Details}\label{app:reward}
\emph{Presence} is the fraction of tagged entities that occur verbatim in the
source note; an annotation with no entities scores $1$. \emph{Completeness} is
entity-level $F_\beta$ with $\beta=2$ under guarded substring matching: a
length guard on containment matches closes a reward-hacking channel in which
very short outputs match arbitrary references.
\emph{Calibration} compares the claimed \texttt{<qc\_status>} with whether the
annotation is complete (Table~\ref{tab:qc-reward}). \emph{Length} gives full
credit up to $1{,}024$ tokens and decays linearly beyond.

\begin{table}[!htb]
  \centering\small
  \setlength{\tabcolsep}{4pt}
  \begin{tabular}{l l c}
    \toprule
    Claimed & Actual & $r_{\mathrm{qc}}$ \\
    \midrule
    complete   & complete   & 1.0 \\
    incomplete & incomplete & 1.0 \\
    incomplete & complete   & 0.5 \\
    complete   & incomplete & 0.0 \\
    (tag missing) & any     & 0.0 \\
    \bottomrule
  \end{tabular}
  \caption{The calibration reward. Claiming completeness after a miss is the
  safety-critical error and scores zero.}
  \label{tab:qc-reward}
\end{table}

\section{Why the Self-Assessment Tag Was Never Emitted}\label{app:qc_calibration}
The \texttt{<qc\_status>} tag is absent from all outputs of the SFT and GRPO
models, including when GRPO is prompted with its own training instruction
block, which states the tag as mandatory, and decoded without a stop string.
The outputs end at \texttt{</phi\_entities>} well within the generation budget:
the policy closes the annotation and stops.

The reward's structure explains this. The supervised targets end at
\texttt{</phi\_entities>}, so the initial policy assigns almost no probability
to any continuation. GRPO standardizes rewards within a group of sampled
completions, and the calibration term is exactly $0$ whenever the tag is
absent. When no completion in a group carries the tag, which is the
overwhelmingly likely case at initialization, the term is constant across the
group, contributes no variance, and vanishes from the standardized advantage.
Its $0.15$ of reward mass was inert throughout training. A reward that is zero
for both a wrong answer and a missing answer gives the model no path from the
second to the first.

Two remedies follow: put the tag in the supervised targets so that GRPO shapes
it instead of having to discover it, or make the verdict the only thing the
model writes. The verifier of Section~\ref{sec:verifier} takes the second
route.

For the same reason, the QC layer does not treat a missing tag as a sign
that a note is unreliable (Section~\ref{sec:qc}).

\section{QC Layer Details}\label{app:qc_ablation}
Repair first tries a strict XML parser and then a recovering one (lxml with
\texttt{recover}), keeping every tag it can read; an output that neither can
read becomes an empty annotation. Pattern detection rejects numeric dates whose
month or day is impossible (``10/70''). The NER fallback maps spaCy's PERSON
label to name and GPE, LOC, FAC, and ORG to place, adds a street-address
pattern, and skips abbreviations (``ECG'', ``CAD''), generic institution words,
and notes that are more than $60\%$ upper case. A note is sent to the NER
fallback if its XML could not be repaired, if pattern detection added three or
more spans, or if the model's own \texttt{<qc\_status>} tag reports
\texttt{incomplete}; the last never happens, because the model never emits the
tag (Appendix~\ref{app:qc_calibration}).

Table~\ref{tab:qc_layers} adds the steps one at a time to the SFT model. On the
benchmark only pattern detection changes anything, adding one entity that is
not \PHI. On MIMIC nearly all of the effect comes from repair: $2.6\%$ of SFT
outputs do not parse, and recovering their tags raises recall by $0.037$ and
lowers precision by $0.129$, since about nine in ten of the recovered values are
not \PHI. Pattern detection adds $0.009$ recall, and the NER fallback runs on
one note. On the $404$-note held-out split of the training corpus the layer
adds $17$ entities, $3$ of them \PHI, and F1 moves from $0.987$ to $0.986$.

\begin{table}[!htb]
\centering\small
\setlength{\tabcolsep}{3pt}
\begin{tabular}{@{}l ccc ccc@{}}
\toprule
 & \multicolumn{3}{c}{Benchmark} & \multicolumn{3}{c}{MIMIC} \\
\cmidrule(lr){2-4}\cmidrule(l){5-7}
Step & P & R & F1 & P & R & F1 \\
\midrule
None              & 0.990 & 0.999 & 0.994 & 0.407 & 0.580 & 0.478 \\
+ Repair          & 0.990 & 0.999 & 0.994 & 0.278 & 0.617 & 0.383 \\
+ Patterns        & 0.989 & 0.999 & 0.994 & 0.280 & 0.626 & 0.387 \\
+ NER fallback    & 0.989 & 0.999 & 0.994 & 0.280 & 0.626 & 0.387 \\
\bottomrule
\end{tabular}
\caption{The QC layer step by step on \model SFT (GRPO is identical on the
benchmark). MIMIC is scored against the human gold standard.}
\label{tab:qc_layers}
\end{table}

\section{Significance Tests}\label{app:significance}
Table~\ref{tab:sig-full} lists paired bootstrap differences ($1{,}000$
resamples over notes, seed 2025) for the comparisons discussed in the paper.

\begin{table*}[!t]
\centering\small
\setlength{\tabcolsep}{2.5pt}
\begin{tabular}{@{}l l c c c c@{}}
\toprule
Test set & Comparison ($B - A$) & $\Delta$ Recall & $\Delta$ Precision & $\Delta$ F1 & $\Delta$ Zero-leak \\
\midrule
Bench. & GRPO $-$ SFT
  & $0.000$ & $0.000$ & $0.000$ & $0.000$ \\
 & & \ci{$0.000$}{$0.000$} & \ci{$0.000$}{$0.000$} & \ci{$0.000$}{$0.000$} & \ci{$0.000$}{$0.000$} \\
Bench. & QC $-$ none, on GRPO
  & $0.000$ & $-0.001$ & $0.000$ & $0.000$ \\
 & & \ci{$0.000$}{$0.000$} & \ci{$-0.003$}{$0.000$} & \ci{$-0.001$}{$0.000$} & \ci{$0.000$}{$0.000$} \\
\midrule
MIMIC & GRPO $-$ SFT
  & $-0.052$ & $+0.112$ & $+0.046$ & $-0.022$ \\
 & & \ci{$-0.083$}{$-0.022$} & \ci{$+0.084$}{$+0.146$} & \ci{$+0.021$}{$+0.074$} & \ci{$-0.029$}{$-0.016$} \\
MIMIC & QC $-$ none, on SFT
  & $+0.046$ & $-0.126$ & $-0.091$ & $+0.005$ \\
 & & \ci{$+0.023$}{$+0.073$} & \ci{$-0.198$}{$-0.033$} & \ci{$-0.164$}{$-0.013$} & \ci{$+0.003$}{$+0.009$} \\
MIMIC & SFT, few-shot $-$ zero-shot
  & $-0.019$ & $+0.155$ & $+0.083$ & $-0.009$ \\
 & & \ci{$-0.052$}{$+0.016$} & \ci{$+0.122$}{$+0.193$} & \ci{$+0.057$}{$+0.113$} & \ci{$-0.017$}{$-0.003$} \\
MIMIC & GRPO, few-shot $-$ zero-shot
  & $+0.079$ & $+0.088$ & $+0.084$ & $+0.027$ \\
 & & \ci{$+0.047$}{$+0.117$} & \ci{$+0.061$}{$+0.116$} & \ci{$+0.060$}{$+0.110$} & \ci{$+0.019$}{$+0.035$} \\
\bottomrule
\end{tabular}
\caption{Paired bootstrap differences with $95\%$ intervals. An interval that
excludes zero marks a significant difference. Because GRPO and SFT have
identical scored predictions on the benchmark, QC on SFT gives the same row as
QC on GRPO. Few-shot uses three hand-written example notes in the
prompt; MIMIC rows are scored against the human gold standard over all
categories.}
\label{tab:sig-full}
\end{table*}

\section{Per-Category Results}\label{app:per-entity}
Tables~\ref{tab:per-entity} and~\ref{tab:mimic-per-entity} give per-category
F1 on the two test sets. A dash marks a category the system has no label for,
which is different from a score of zero (category attempted, never matched).

\begin{table*}[!t]
  \centering\small
  \setlength{\tabcolsep}{3pt}
  \begin{tabular}{@{}l c c c c c c c c c c@{}}
    \toprule
    Category & Presidio & GLiNER & RoBERTa & BERT & OpenMed & Stanford & spaCy & OPF & Base & \model \\
             &          &        & -i2b2   & -i2b2 & -PII   & -deid    &       &     & 0-shot & SFT \\
    \midrule
    Name    & 0.990 & 0.842 & 0.647 & 0.616 & 0.646 & 0.983 & 0.982 & 0.986 & 0.352 & \textbf{0.996} \\
    Age     & --    & 0.756 & \textbf{0.987} & \textbf{0.987} & \textbf{0.987} & -- & -- & -- & 0.280 & \textbf{0.987} \\
    Date    & 0.452 & 0.973 & 0.039 & 0.024 & 0.004 & 0.990 & 0.446 & 0.986 & 0.110 & \textbf{0.998} \\
    Place   & --    & 0.868 & 0.000 & 0.000 & 0.000 & 0.221 & 0.629 & --    & 0.000 & \textbf{0.972} \\
    Address & 0.000 & 0.000 & 0.000 & 0.000 & 0.000 & --    & 0.000 & 0.938 & 0.128 & \textbf{1.000} \\
    Phone   & 0.156 & 0.958 & 0.103 & 0.000 & 0.054 & 0.000 & --    & 0.978 & 0.045 & \textbf{1.000} \\
    Email   & \textbf{1.000} & 0.316 & 0.660 & 0.000 & 0.984 & -- & -- & \textbf{1.000} & 0.173 & \textbf{1.000} \\
    ID      & --    & 0.730 & 0.865 & 0.000 & 0.959 & 0.000 & --    & 0.959 & 0.096 & \textbf{0.984} \\
    URL     & 0.674 & 0.352 & --    & --    & 0.267 & --    & --    & 0.114 & 0.134 & \textbf{1.000} \\
    Device  & --    & 0.370 & --    & --    & 0.000 & 0.000 & --    & --    & 0.000 & \textbf{0.923} \\
    \bottomrule
  \end{tabular}
  \caption{Per-category F1 on the in-domain benchmark. GRPO's predictions are
  identical to SFT's.}
  \label{tab:per-entity}
\end{table*}

\begin{table*}[!t]
  \centering\small
  \setlength{\tabcolsep}{3pt}
  \begin{tabular}{@{}l c c c c c c c c c c c@{}}
    \toprule
    Category & RoBERTa & Stanford & BERT & GLiNER & OPF & OpenMed & Presidio & spaCy & Base & \model & \model \\
             & -i2b2   & -deid    & -i2b2 &       &     & -PII    &          &       & 0-shot & SFT & GRPO \\
    \midrule
    Name    & \textbf{0.854} & 0.768 & 0.741 & 0.597 & 0.574 & 0.638 & 0.226 & 0.231 & 0.019 & 0.541 & 0.598 \\
    Date    & \textbf{0.745} & 0.090 & 0.172 & 0.547 & 0.205 & 0.269 & 0.066 & 0.058 & 0.030 & 0.509 & 0.589 \\
    Place   & 0.481 & \textbf{0.710} & 0.396 & 0.370 & 0.000 & 0.008 & 0.000 & 0.015 & 0.000 & 0.456 & 0.422 \\
    Phone   & 0.561 & 0.389 & 0.541 & 0.302 & 0.468 & \textbf{0.604} & 0.349 & 0.000 & 0.000 & 0.175 & 0.267 \\
    Age     & 0.000 & 0.000 & 0.036 & 0.032 & 0.000 & 0.047 & 0.000 & 0.000 & 0.000 & \textbf{0.048} & 0.045 \\
    ID      & 0.026 & \textbf{0.105} & 0.100 & 0.000 & 0.000 & 0.041 & 0.000 & 0.000 & 0.000 & 0.000 & 0.000 \\
    Address & 0.000 & --    & 0.000 & 0.000 & 0.000 & 0.000 & 0.000 & 0.000 & 0.000 & 0.000 & 0.000 \\
    \bottomrule
  \end{tabular}
  \caption{Per-category F1 on MIMIC against the human gold standard. Ages are
  near zero for every system that predicts them because MIMIC annotates only
  ages of $90$ and above ($4$ gold spans), so almost every predicted age is a
  false positive.}
  \label{tab:mimic-per-entity}
\end{table*}

\section{MIMIC Analysis}\label{app:mimic}

\paragraph{Reference annotation.}
The corpus is distributed with the notes, a human gold standard, and the
output of a rule-based de-identification tool. One of the distributed annotation files is in fact that tool's output
($99.5\%$ of its spans coincide with the tool's detections) and is easy to
mistake for the reference; about $26\%$ of its spans are not \PHI. Table~\ref{tab:mimic-ref} shows how much
the choice matters. Every competitive system scores higher against the human
gold standard, by $0.03$ to $0.13$ F1, and the ordering of the middle of the
table changes.

\begin{table}[!htb]
\centering\small
\setlength{\tabcolsep}{4pt}
\begin{tabular}{@{}l c c c@{}}
\toprule
System & Tool ref. & Human gold & Untyped \\
\midrule
RoBERTa-i2b2    & 0.571 & 0.675 & 0.708 \\
Stanford-deid   & 0.435 & 0.529 & 0.535 \\
BERT-i2b2       & 0.398 & 0.525 & 0.560 \\
Ours, GRPO      & 0.442 & 0.523 & 0.542 \\
Ours, SFT       & 0.423 & 0.478 & 0.500 \\
GLiNER          & 0.410 & 0.475 & 0.486 \\
OPF             & 0.350 & 0.410 & 0.421 \\
OpenMed-PII     & 0.311 & 0.342 & 0.372 \\
Presidio        & 0.116 & 0.115 & 0.135 \\
spaCy           & 0.064 & 0.063 & 0.079 \\
\bottomrule
\end{tabular}
\caption{MIMIC F1 against the rule-based tool's output, against the human gold
standard, and against the human gold standard ignoring entity type. ``Ours''
is \model.}
\label{tab:mimic-ref}
\end{table}

\paragraph{Safe Harbor classification.}
We classified every gold span against the HIPAA Safe Harbor list with rules
over the gold type and its context (Table~\ref{tab:hipaa}). This is not a
legal determination. Less than half of the annotated spans are Safe Harbor
identifiers; a third are clinician names, which many institutions remove by
policy but which are not identifiers of the patient.

\begin{table}[!htb]
\centering\small
\setlength{\tabcolsep}{3pt}
\begin{tabular}{@{}l r r r@{}}
\toprule
Class of gold span & Spans & \multicolumn{2}{c}{Missed} \\
                   &       & Ours & RoB. \\
\midrule
\multicolumn{4}{@{}l}{\emph{Safe Harbor \PHI} ($823$)} \\
\quad date (month/day)            & 482 & 113 & 57 \\
\quad patient or relative name    & 231 & 87  & 40 \\
\quad city, town, region, street  & 56  & 49  & 3 \\
\quad patient or family phone     & 42  & 18  & 11 \\
\quad other (employer, age $>89$) & 12 & 8 & 9 \\
\midrule
\multicolumn{4}{@{}l}{\emph{Not Safe Harbor} ($734$)} \\
\quad clinician name              & 593 & 203 & 94 \\
\quad ward or unit name           & 74  & 69  & 64 \\
\quad year only                   & 46  & 20  & 12 \\
\quad other (pager, vendor, \ldots) & 21 & 10 & 10 \\
\midrule
\multicolumn{4}{@{}l}{\emph{Unclear} ($222$)} \\
\quad hospital or facility name   & 219 & 116 & 23 \\
\quad other                       & 3   & 2   & 0 \\
\bottomrule
\end{tabular}
\caption{MIMIC gold spans by Safe Harbor status, with the number left
uncovered by any prediction, of any type, of \model SFT (Ours) and of
RoBERTa-i2b2 (RoB.). The table counts the $1{,}779$ spans as annotated; two
overlapping Location spans are merged into one for scoring, giving the
$1{,}778$ spans used elsewhere.}
\label{tab:hipaa}
\end{table}

\paragraph{Why names and dates are missed.}
Table~\ref{tab:miss} summarizes the analysis behind
Section~\ref{sec:mimic}. Its scope is the $713$ gold spans that are patient or
relative names, patient initials, or dates, in $373$ notes; the SFT model
misses $200$. Forty of these misses are spans the model did tag in an output
that a later malformed tag made unparseable; with such artifacts removed,
recall on these spans is about $0.78$ instead of $0.72$.

\begin{table}[!htb]
\centering\small
\setlength{\tabcolsep}{3pt}
\begin{tabular}{@{}p{4.7cm} r@{}}
\toprule
Measurement & Value \\
\midrule
\multicolumn{2}{@{}l}{\emph{Whole-note failures}} \\
Misses in notes with empty or unparseable output & 155 / 200 \\
\quad output parseable but empty & 103 \\
\quad output unparseable & 52 \\
Benchmark notes with such output & 0 / 100 \\
Miss rate in notes with normal output & 8.1\% \\
\quad first vs.\ third quarter of the note & 4.3\% vs.\ 14.5\% \\
\midrule
\multicolumn{2}{@{}l}{\emph{Coverage: training / benchmark / MIMIC}} \\
\PHI per $1{,}000$ characters & 10.8 / 14.9 / 0.92 \\
Notes with $\le 2$ \PHI mentions & 2.5\% / 1.0\% / -- \\
Mostly upper-case notes & 0\% / -- / 29.8\% \\
Bare month/day dates & 15.0\% / 0\% / 81.1\% \\
Names typed as relative & 0 of 151{,}088 / -- / -- \\
\midrule
\multicolumn{2}{@{}l}{\emph{MIMIC name miss rate by surface form}} \\
Title case / lower case / upper case & 17\% / 60\% / 69\% \\
Relative vs.\ patient name & 40\% vs.\ 28\% \\
\midrule
\multicolumn{2}{@{}l}{\emph{Controlled edits of missed cases ($n{=}25$ each)}} \\
Prepend ``Patient: \emph{name}'' header & 76\% recovered \\
Keep only the target sentence & 68\% recovered \\
Rewrite date as ``Month D, YYYY'' & 28\% recovered \\
Title-case the name & 20\% recovered \\
Sentence-case an upper-case note & 20\% recovered \\
\bottomrule
\end{tabular}
\caption{Analysis of missed Safe Harbor names and dates on MIMIC (SFT
model). Applying the reverse edits to benchmark notes (lower-case names,
year-less dates, upper-case notes, ICU shorthand) leaves benchmark recall
between $99.5\%$ and $100\%$.}
\label{tab:miss}
\end{table}

\paragraph{Input case.}
In a separate run served with vLLM, lower-casing every MIMIC note lowers SFT
recall from $0.572$ to $0.442$, while lower-casing only the mostly upper-case
notes raises precision ($0.415$ to $0.453$) at a small recall cost ($0.572$ to
$0.542$). On the benchmark, lower-casing changes F1 by less than $0.005$. The
model uses capitalization as a cue for names, and the cue is unreliable in
both directions on nursing notes.

\section{Robustness to Synthetic Perturbation}\label{app:robustness}
We perturb the benchmark's inline-tagged gold text and re-derive the note and
its reference, so the two stay consistent (Table~\ref{tab:robustness}). The
\emph{surface} profile rotates \PHI surface forms (date formats, phone
separators, name spellings, identifier formats). The \emph{register} profile
applies document-level noise: upper-casing, joined paragraphs, dropped
articles and prepositions, and semicolon-separated clauses. Because the
reference is re-derived from inline tags, the clean row of this harness
differs slightly from Table~\ref{tab:main}.

\begin{table}[!htb]
\centering\small
\setlength{\tabcolsep}{4pt}
\begin{tabular}{@{}l c c c c@{}}
\toprule
Profile & P & R & F1 & Zero-leak \\
\midrule
Clean    & 0.992 & 0.996 & 0.994 & 0.96 \\
Surface  & 0.992 & 0.996 & 0.994 & 0.96 \\
Register & 0.986 & 0.983 & 0.985 & 0.85 \\
Both     & 0.984 & 0.955 & 0.969 & 0.75 \\
\bottomrule
\end{tabular}
\caption{\model (SFT) under synthetic perturbations of the benchmark.}
\label{tab:robustness}
\end{table}

\begin{table*}[!t]
\centering\small
\begin{tabular}{l l c c c c c c c}
\toprule
Test set & Verifier & Base rate & AUROC & ECE & Flagged & Caught & Flag precision & Residual \\
\midrule
Synthetic val ($800$)  & zero-shot  & 0.301 & 0.537 & 0.024 & 0.105 & 0.154 & 0.441 & 0.285 \\
                       & SFT        &       & 0.813 & 0.013 & 0.204 & 0.556 & 0.822 & 0.168 \\
                       & SFT + GRPO &       & 0.809 & 0.134 & 0.163 & 0.506 & 0.939 & 0.178 \\
\midrule
Benchmark ($1{,}100$)  & zero-shot  & 0.795 & 0.600 & 0.531 & 0.160 & 0.199 & 0.989 & 0.758 \\
                       & SFT        &       & 0.987 & 0.139 & 0.746 & 0.931 & 0.991 & 0.215 \\
                       & SFT + GRPO &       & 0.989 & 0.213 & 0.574 & 0.722 & 1.000 & 0.518 \\
\midrule
MIMIC ($26{,}774$)     & zero-shot  & 0.178 & 0.658 & 0.027 & 0.024 & 0.036 & 0.275 & 0.175 \\
                       & SFT        &       & 0.661 & 0.087 & 0.175 & 0.280 & 0.285 & 0.155 \\
                       & SFT + GRPO &       & 0.661 & 0.162 & 0.085 & 0.147 & 0.309 & 0.165 \\
\bottomrule
\end{tabular}
\caption{Verifier results, pooled over systems. Base rate: share of
(note, annotation) pairs that are incomplete. Flagged: share sent to review.
Caught: share of incomplete pairs that are flagged. Residual: share of
unflagged pairs that are incomplete.}
\label{tab:verifier-full}
\end{table*}

\begin{table*}[!t]
\centering\small
\begin{tabular}{l l l l r r r}
\toprule
System & Device & Runtime & Precision & Median ms & p90 ms & Peak RSS (MB) \\
\midrule
spaCy            & CPU & spaCy        & fp32 & 35 & 47 & 1{,}319 \\
Presidio         & CPU & spaCy/regex  & fp32 & 51 & 70 & 1{,}536 \\
BERT-i2b2        & CPU & transformers & fp32 & 119 & 187 & 1{,}233 \\
Stanford-deid    & CPU & transformers & fp32 & 124 & 166 & 1{,}222 \\
GLiNER           & CPU & transformers & fp32 & 192 & 294 & 2{,}418 \\
RoBERTa-i2b2     & CPU & transformers & fp32 & 335 & 466 & 2{,}118 \\
OpenMed-PII      & CPU & transformers & fp32 & 476 & 624 & 2{,}285 \\
OPF              & CPU & torch        & fp32 & 12{,}413 & 19{,}875 & 4{,}778 \\
\model (SFT) + prompt cache & CPU & llama.cpp & Q4\_K\_M & 3{,}931 & 7{,}711 & 615 \\
\model (SFT) + prompt cache & CPU & llama.cpp & Q8\_0    & 4{,}494 & 8{,}953 & 616 \\
\model (SFT)     & CPU & llama.cpp    & Q4\_K\_M & 5{,}914 & 9{,}477 & 615 \\
\model (SFT)     & CPU & llama.cpp    & Q8\_0    & 6{,}211 & 10{,}527 & 617 \\
\model (GRPO) & CPU & llama.cpp    & Q4\_K\_M & 5{,}808 & 9{,}446 & 615 \\
\model (SFT)     & CPU & transformers & fp32 & 29{,}484 & 51{,}388 & 4{,}248 \\
\model (GRPO) & CPU & transformers & fp32 & 29{,}085 & 51{,}095 & 4{,}265 \\
Qwen3-0.6B-Base  & CPU & transformers & fp32 & 150{,}609 & 153{,}169 & 4{,}238 \\
\midrule
Stanford-deid    & GPU & transformers & bf16 & 8 & 11 & 1{,}295 \\
BERT-i2b2        & GPU & transformers & bf16 & 9 & 19 & 1{,}289 \\
RoBERTa-i2b2     & GPU & transformers & bf16 & 14 & 19 & 2{,}244 \\
GLiNER           & GPU & transformers & bf16 & 18 & 127 & 2{,}420 \\
OPF              & GPU & torch        & bf16 & 24 & 31 & 1{,}994 \\
OpenMed-PII      & GPU & transformers & bf16 & 26 & 28 & 2{,}637 \\
\model (SFT)     & GPU & vLLM         & Q4\_K\_M & 441 & 765 & -- \\
\model (SFT)     & GPU & vLLM         & FP8  & 442 & 796 & -- \\
\model (SFT)     & GPU & vLLM         & bf16 & 511 & 931 & -- \\
\model (GRPO) & GPU & vLLM         & bf16 & 509 & 926 & -- \\
\model (SFT)     & GPU & transformers & bf16 & 3{,}990 & 6{,}689 & 2{,}073 \\
\model (GRPO) & GPU & transformers & bf16 & 3{,}982 & 6{,}599 & 2{,}090 \\
Qwen3-0.6B-Base  & GPU & transformers & bf16 & 18{,}630 & 19{,}105 & 2{,}051 \\
\bottomrule
\end{tabular}
\caption{Single-note latency (median and 90th percentile over $20$ notes after
warm-up) and peak resident memory. GPU: NVIDIA L40S. CPU: AMD EPYC 9354
limited to $8$ threads on a shared server. With $100$ concurrent requests,
vLLM serves $31$ notes per second at bf16 and $17$ at Q4\_K\_M. Quantized
variants score $0.994$ F1 on the benchmark with a \zeroleak rate of $0.98$
(Q4\_K\_M) or $0.99$ (Q8\_0, FP8).}
\label{tab:latency-full}
\end{table*}

Surface variation has no effect, which is what the augmentation was meant to
achieve. Register noise costs $0.013$ recall and $0.11$ \zeroleak, and the
combination $0.041$ and $0.21$. These effects are real but far smaller than
the drop on MIMIC, which is consistent with \PHI density and note structure,
and not casing or telegraphic style alone, driving that drop.

\section{Verifier Details}\label{app:verifier}
Table~\ref{tab:verifier-full} reports the verifier as a review filter at the
threshold $P(\texttt{incomplete})>1/3$, which is the reward-optimal threshold
given that over-caution costs half as much as overconfidence. The threshold was fixed before looking at either test set. The benchmark pool
holds the outputs of all eleven systems of Table~\ref{tab:main}. Four of them
(BERT-i2b2, GLiNER, Presidio, and RoBERTa-i2b2) also produced some of the
verifier's training candidates, on different, synthetic training notes. On the
benchmark, $198$ of the $226$ complete pairs are outputs of \model (SFT, with
and without QC), so the pooled AUROC mostly reflects how well the verifier
accepts \model's complete outputs. On the nine other systems alone, with $28$
complete pairs among $900$, it is $0.938$ for the SFT adapter, $0.941$ after
GRPO, and $0.766$ zero-shot. OPF's outputs are the hardest: $12$ of $100$ are
complete, and the adapter separates them from its incomplete ones at $0.789$
AUROC against $0.819$ zero-shot.

On the benchmark the SFT adapter cuts the leak rate among released
annotations from $0.795$ to $0.215$ while reviewing $75\%$ of them. A
threshold chosen to catch $95\%$ of incomplete annotations on synthetic
validation catches $99.9\%$ on the benchmark but flags $95.5\%$ of all pairs,
which is of little practical use; the validation distribution of perturbed
gold annotations is easier to pass than real system output. On \model's own
MIMIC outputs the SFT adapter reaches $0.688$ AUROC against $0.667$ zero-shot.

\section{Full Latency Results}\label{app:latency}
Table~\ref{tab:latency-full} gives all latency measurements. Each system runs
in a fresh process with the device fixed before start-up; the device it ran on
is read back and verified. The zero-shot base model is slow because it never
emits the closing tag and runs to the $1{,}024$-token limit on every note. The
median prompt is about $1{,}060$ tokens, of which the fixed instruction block
is $867$, which is why caching that prefix helps on CPU.

\section{Baseline Details}\label{app:baselines}
All baselines run locally with default thresholds. Their native labels are
mapped to our ten types and serialized into the \texttt{<phi\_entities>}
schema before scoring. Table~\ref{tab:baselines} lists the checkpoints.

\begin{table}[!htb]
\centering\small
\setlength{\tabcolsep}{3pt}
\begin{tabular}{@{}l p{4.6cm}@{}}
\toprule
System & Checkpoint or configuration \\
\midrule
Presidio      & Microsoft Presidio analyzer, default recognizers \\
spaCy         & \texttt{en\_core\_web\_lg} NER \\
GLiNER        & \texttt{urchade/\allowbreak gliner\_mediumv2.1}, prompted with our type names \\
Stanford-deid & \texttt{StanfordAIMI/\allowbreak stanford-deidentifier-base} \\
RoBERTa-i2b2  & \texttt{obi/\allowbreak deid\_roberta\_i2b2} \\
BERT-i2b2     & \texttt{obi/\allowbreak deid\_bert\_i2b2} \\
OpenMed-PII   & \texttt{OpenMed/\allowbreak OpenMed-PII-\allowbreak SuperClinical-\allowbreak Large-434M-v1} \\
OPF           & OpenAI Privacy Filter, run from its released source \\
Base 0-shot   & \texttt{Qwen/\allowbreak Qwen3-0.6B-Base} with the \model prompt \\
\bottomrule
\end{tabular}
\caption{Baseline systems.}
\label{tab:baselines}
\end{table}

\end{document}